\documentclass{article}

\usepackage[nonatbib,final]{neurips_2024_camera}

\usepackage{amsmath,amssymb,amsfonts}
\usepackage{bbm}
\usepackage{algorithmic}
\usepackage{graphicx}
\usepackage{textcomp}

\usepackage{listings} 
\usepackage{hyperref}

\usepackage{import}

\usepackage{svg}

\usepackage{booktabs}
\usepackage{xcolor}
\usepackage{colortbl}
\usepackage{rotating}
\usepackage{multirow}
\usepackage[normalem]{ulem} 

\usepackage{bbm}
\usepackage[sorting=none]{biblatex} 
\usepackage[utf8]{inputenc} 
\usepackage[T1]{fontenc}    
\usepackage{hyperref}       
\usepackage{url}            
\usepackage{booktabs}       
\usepackage{amsfonts}       
\usepackage{nicefrac}       
\usepackage{microtype}      
\usepackage{xcolor}         

\title{SynEHRgy: Synthesizing Mixed-Type Structured Electronic Health Records using Decoder-Only Transformers}

\author{
  Hojjat Karami \\
  EPFL\\
  \texttt{hojjat.karami@epfl.ch}
  \And
  David Atienza \\
  EPFL\\
  \texttt{david.atienza@epfl.ch}
  \And
  Anisoara Ionescu \\
  EPFL\\
  \texttt{anisoara.ionescu@epfl.ch}
}

\begin{document}

\maketitle

\begin{abstract}
  Generating synthetic Electronic Health Records (EHRs) offers significant potential for data augmentation, privacy-preserving data sharing, and improving machine learning model training. We propose a novel tokenization strategy tailored for structured EHR data, which encompasses diverse data types such as covariates, ICD codes, and irregularly sampled time series. Using a GPT-like decoder-only transformer model, we demonstrate the generation of high-quality synthetic EHRs. Our approach is evaluated using the MIMIC-III dataset, and we benchmark the fidelity, utility, and privacy of the generated data against state-of-the-art models.

\end{abstract}

\section{Introduction}

Electronic Health Records (EHRs) are comprehensive digital repositories of patient health information, encompassing a wide range of data types. In in-patient settings, EHRs include structured data such as demographics and International Classification of Diseases (ICD) codes, time series data (e.g., vital signs and lab results), and unstructured data (e.g., clinical notes and radiology images). These records provide an invaluable resource for developing machine learning models, offering rich datasets that can be harnessed to enhance patient care, predict outcomes, and support clinical decision-making.


Access to real-world EHRs is often limited due to privacy regulations such as HIPAA and GDPR, as well as technical challenges and a lack of incentives for data sharing. Even pseudo-anonymized data remains susceptible to re-identification by malicious actors. Techniques like federated learning \cite{riekeFutureDigitalHealth2020} and differential privacy \cite{khalidPrivacypreservingArtificialIntelligence2023} are being explored to mitigate these risks. Synthetic data provides a promising alternative by replicating the statistical properties of real data without being tied to actual individuals, thereby circumventing privacy regulations. It can be securely shared among stakeholders, facilitating hypothesis generation, AI model pre-training, and data exploration without exposing sensitive information. This approach not only alleviates privacy concerns but also reduces the effort involved in accessing real data, allowing researchers to focus on validating more probable hypotheses before turning to original datasets for confirmation.


Generating synthetic EHR data presents several challenges. EHRs comprise multiple data types with distinct characteristics. For instance, clinical events like ICD codes are high-dimensional and sequential, while time series data are relatively low-dimensional but feature irregular sampling and significant missingness, which is often informative and not random \cite{ghassemiReviewChallengesOpportunities2020,tanInformativeMissingnessWhat2023}. Additionally, patients may have multiple visits or admissions, with data across visits being correlated. While many studies focus on generating single data types, such as discrete ICD codes (\cite{choiGeneratingMultilabelDiscrete2017a,torfiCorGANCorrelationCapturingConvolutional2020,biswalEVAGeneratingLongitudinal2020,wangPromptEHRConditionalElectronic2022}) or time series data (\cite{peiGeneratingRealWorldTime2021a,tianFastReliableGeneration2023,karamiTimEHRImagebasedTime2024}), few addresses the generation of multiple data types across multiple visits simultaneously (\cite{theodorouSynthesizeHighdimensionalLongitudinal2023}).


In the era of Large Language Models (LLMs), many have been adapted to the clinical domain for tasks such as diagnosis prediction \cite{liBEHRTTransformerElectronic2019}, text embeddings \cite{huangClinicalBERTModelingClinical2020b}, and medical reasoning \cite{singhalExpertLevelMedicalQuestion2023, chenMEDITRON70BScalingMedical2023}. While these models can interpret structured data and numerical values presented in text format, they are not directly applicable for generating synthetic structured EHR data. This limitation arises because LLMs are typically not trained on large structured EHR datasets, particularly those containing complex numerical data.




In this work, we present \textbf{SynEHRgy}, a framework designed to \textbf{Syn}thesize mixed-type structured \textbf{EHR}s, including covariates, ICD codes, and irregularly-sampled time series across multiple patient visits. We introduce a novel tokenization strategy for structured EHR data, particularly for numerical variables, allowing decoder-only transformer models to capture underlying data patterns within a causal language modeling framework and generate high-quality synthetic EHRs. Using a small GPT model, we demonstrate the efficacy of our approach on the MIMIC-III dataset \cite{johnsonMIMICIIIFreelyAccessible2016}, comparing the generated data's quality against state-of-the-art models in terms of fidelity, utility, and privacy. The key contributions of this work are as follows:

\begin{itemize}
  \item We propose a novel tokenization strategy for mixed-type structured EHR data (covariates, ICD codes, irregularly-sampled time series) across multiple visits.
  \item We empirically demonstrate that a GPT-like decoder-only transformer model can generate high-quality structured EHR data.
  \item We conduct a comprehensive evaluation of the proposed methodology on the MIMIC-III dataset, comparing the generated data's quality to state-of-the-art models in terms of fidelity, utility, and privacy.
  \item Our framework shows exceptional performance in generating irregularly-sampled time series data, a challenging task due to high missingness and irregular time points. \end{itemize}

\section{Related Works}




Generating clinical events, particularly ICD codes, has garnered significant research interest in recent years. In contrast, generating clinical time series presents a greater challenge due to irregularly-sampled time points and substantial missingness.

\paragraph{ICD Code Generation}

A substantial body of work focuses on generating ICD codes using various frameworks. Generative Adversarial Networks (GANs) have been employed for this purpose (\cite{baowalySynthesizingElectronicHealth2019, torfiCorGANCorrelationCapturingConvolutional2020, zhangSynTEGFrameworkTemporal2021, luMultiLabelClinicalTimeSeries2023}), as have Variational Autoencoders (VAEs) (\cite{nikolentzosSyntheticElectronicHealth2023}) and diffusion models (\cite{heMedDiffGeneratingElectronic2023, ceritliSynthesizingMixedtypeElectronic2023, yuanEHRDiffExploringRealistic2024, zhongSynthesizingMultimodalElectronic2024}). Additionally, Large Language Models (LLMs) have been explored due to the sequential nature of ICD codes and the feasibility of straightforward tokenization. For instance, PromptEHR \cite{wangPromptEHRConditionalElectronic2022} employs an encoder-decoder transformer model with prompt learning to generate ICD codes conditioned on numerical and categorical demographic features. HALO \cite{theodorouSynthesizeHighdimensionalLongitudinal2023} utilizes a transformer encoder coupled with a linear autoregressive module for generating ICD codes at both the visit and code levels. CEHR-GPT \cite{pangCEHRGPTGeneratingElectronic2024} leverages GPT to produce sequences of visits based on demographic prompts and visit time intervals.

\paragraph{Time Series Generation}

Research on time series generation has predominantly focused on regularly-sampled time series, which are more common across various domains. Techniques such as Generative Adversarial Networks (GANs) have been widely applied (\cite{estebanRealvaluedMedicalTime2017a, yoonTimeseriesGenerativeAdversarial2019a, linUsingGANsSharing2020, liGeneratingSyntheticMixedtype2023, yoonEHRSafeGeneratingHighfidelity2023}), as well as diffusion models (\cite{yuanDiffusionTSInterpretableDiffusion2023, qianTimeLDMLatentDiffusion2024, suhTimeAutoDiffCombiningAutoencoder}). However, generating irregularly-sampled time series is less explored. Notable efforts in this area include RTSGAN \cite{peiGeneratingRealWorldTime2021a}, EHR-Safe \cite{yoonEHRSafeGeneratingHighfidelity2023}, and TimEHR \cite{karamiTimEHRImagebasedTime2024} using GANs, and TS-Diffusion \cite{liTSDiffusionGeneratingHighly2023} and TimeDiff \cite{tianFastReliableGeneration2023} using diffusion models. Additionally, research on leveraging language models for time series generation remains limited. While pre-trained LLMs have demonstrated effectiveness with numerical values in text format—e.g., LIFT \cite{dinhLIFTLanguageInterfacedFineTuning2022} and GReaT \cite{borisovLanguageModelsAre2023} excel in tabular data generation—they primarily treat numerical values as text. Some approaches propose defining specific numerical tokens and training LLMs from scratch. For example, CodeAR \cite{leeLeveragingVQVAETokenization2024} utilizes a Vector Quantized Autoencoder (VQ-VAE) for codebook creation, followed by an LSTM with next-token prediction loss. Chronos \cite{ansariChronosLearningLanguage2024} discretizes numerical values into bins for time series forecasting using encoder-decoder transformers, a strategy also employed by HALO \cite{theodorouSynthesizeHighdimensionalLongitudinal2023} for clinical time series and by \cite{grigorasSyntheticTimeSeries2024} for financial time series. To the best of our knowledge, no existing work, other than HALO \cite{theodorouSynthesizeHighdimensionalLongitudinal2023}, explores the generation of irregularly-sampled time series using LLMs.





\section{Methodology}

\subsection{Problem Formulation}

\begin{figure}[t]
  \centering
  \includegraphics[width=0.95\textwidth]{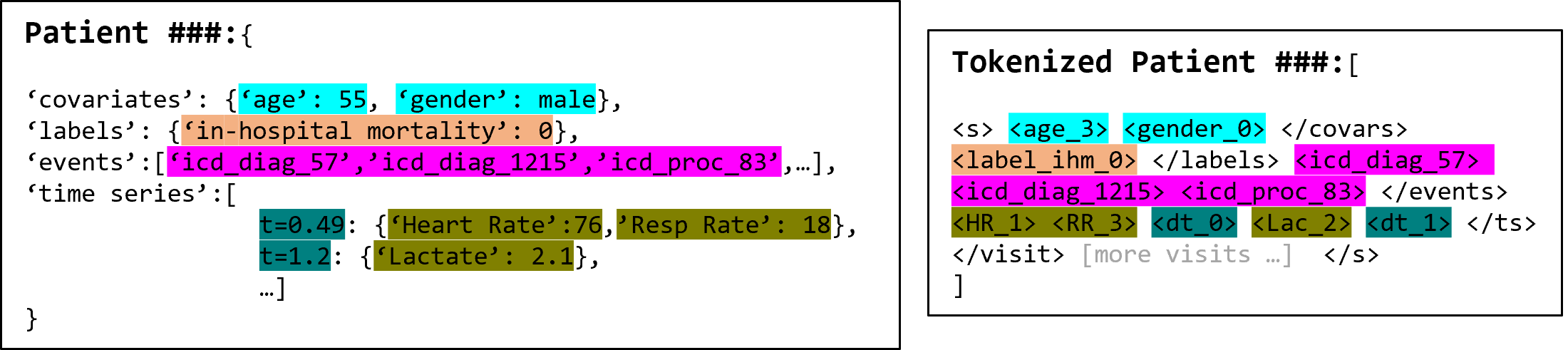}
  \caption{Left: Example of patient's data. Right: Tokenized patient's data}
  \label{fig:pat-sample}
\end{figure}


An EHR dataset can be characterized as $\mathcal{D}=\{P_i\}_{i=1}^{N}$, where $N$ denotes the number of patients and $P_i$ represents the data for the $i$-th patient. Each patient may have one or more visits, denoted as $P_i=\{V_{ij}\}_{j=1}^{|V_i|}$, with each visit represented as $V_{j}=\{c_{j}, y_{j}, E_{j}, T_{j}\}$. Here, $c_{j}$ denotes the set of covariates (e.g., demographics), $y_{j}$ represents the set of clinical outcomes (e.g., in-hospital mortality), $E_{j}=\{e_{m}\}_{m=1}^{|E_{j}|}$ is a sequence of clinical events such as ICD codes, and $T_{j}$ is the time series data, including vital signs and laboratory variables. The time series data for each visit is represented as $T_{j}={\{t_{k}, \{(n_{kp}, x_{kp})\}_{p=1}^{|L_{k}|}\}}_{k=1}^{|T_{j}|}$, where $|T_{j}|$ denotes the total number of measurement times, and $|L_{k}|$ indicates the number of available data points per measurement. In this representation, $t_{k} \in \mathbb{R}{\geq 0}$ is the timestamp of the $k$-th measurement, while $n_{kp}$ and $x_{kp}$ are the variable name and value pairs. An example of patient data is illustrated on the left side of \autoref{fig:pat-sample}.

\subsection{Tokenization}
\label{sec:token}

For numerical values, we employ uniform quantization to discretize numerical variables into equal-sized bins, similar to Chronos \cite{ansariChronosLearningLanguage2024}. Each bin is assigned a unique token. For instance, the token <HR\textunderscore0> represents heart rate values in the range of 55-66 beats per minute. For timestamps, we compute the time intervals between consecutive measurements and discretize these intervals into bins. For non-numerical values such as ICD codes and categorical variables, each code or category is assigned a unique token.

We also use special tokens to delineate different sections of the data: <s> and </s> for the start and end of a sequence, respectively; </covars>, </labels>, </ts>, and </visit> to signify the end of covariates, labels, time series, and admissions. The padding token (<PAD>) is employed to ensure that all tokenized sequences are of uniform length. If a patient has multiple admissions, the data for new admissions is appended after the token </visit>. An example of tokenized patient data is illustrated on the right side of \autoref{fig:pat-sample}.

This tokenization strategy offers several advantages: First, it is scalable with respect to the number of variables. For instance, when integrating a new dataset with additional variables, we simply update the tokenization dictionary by incorporating new tokens. Second, numerical tokenization is more efficient than treating numerical values as text. For example, the GPT-2 tokenizer would generate seven tokens for 'Lactate:2.5', whereas our method generates only one token.

\subsection{Model Architecture}

Decoder-only Transformer models, such as the ones used in language modeling (e.g., GPT \cite{radfordLanguageModelsAre2019}), typically employ a causal language modeling (CLM) objective as their loss function. This objective is also known as autoregressive language modeling. The core idea is to predict the next token in a sequence given all previous tokens. The model generates a probability distribution over the vocabulary for the next token, and the loss function measures the discrepancy between the predicted distribution and the actual next token. For an tokenized patient sequence \( x = [x_1, x_2, \dots, x_T] \), the model predicts the probability distribution over the vocabulary for each token in the sequence, conditioned on all previous tokens:
\[
  p(x_t \mid x_1, x_2, \dots, x_{t-1})
\]

The loss for a single token prediction is typically computed using cross-entropy between the predicted distribution and the one-hot encoded true next token. For the entire sequence, the loss function is:
\[
  \mathcal{L} = - \sum_{t=1}^{T} \log p(x_t \mid x_1, x_2, \dots, x_{t-1})
\]

\subsection{EHR Data Generation}


The generation of EHR data follows a process similar to text generation. We initiate the generation with the start token (<s>) and continue generating tokens sequentially until the end token (</s>) is produced. For the de-tokenization of numerical tokens, we apply uniform sampling within the specified bin range to recover the original numerical values. This approach ensures that the generated values are representative of the underlying distribution of the data. The de-tokenization of ICD codes and categorical variables is straightforward, as each token directly maps to a specific code or category.

This method offers several advantages: First, it leverages the well-established text generation framework, which allows for seamless integration with existing models and techniques. Second, uniform sampling for numerical values maintains the integrity of the data distribution, providing a realistic representation of the original values. Lastly, the straightforward de-tokenization of categorical variables ensures simplicity and efficiency in generating accurate categorical data.

\section{Experiments}
\subsection{Dataset}

We utilize the MIMIC-III dataset for our experiments \cite{johnsonMIMICIIIFreelyAccessible2016}. Specifically, we use the preprocessing pipeline provided by \cite{harutyunyanMultitaskLearningBenchmarking2019} to extract data from approximately 42,000 patients with multiple hospital visits. The dataset is divided into training, validation, and test sets using a 70-15-15 split. For each patient, we include age and gender as covariates. Additionally, for each visit, we select 25 phenotype labels along with an in-hospital mortality label. We include ICD diagnosis and procedure codes with a frequency greater than five across the entire dataset, resulting in 4,656 unique ICD codes. We also select 41 time series from vital signs and laboratory variables, including five categorical time series. After tokenization, we have a total of 5,127 unique tokens, with the training split consisting of 29.6 million tokens. Each patient’s sequence starts with the <s> token and ends with the </s> token, comprising demographics, ICD codes, and time series tokens, as explained in Section \ref{sec:token}.

\subsection{Evaluation Metrics}




We evaluate the synthetic data based on \textbf{Fidelity}, \textbf{Utility}, and \textbf{Privacy}. Since all of our baselines, except HALO, do not generate multiple data types, we assess our metrics separately for each data type. Additionally, we use the test split as a reference for high-quality synthetic data to evaluate these metrics.

\paragraph{Fidelity}

For ICD codes, similar to \cite{theodorouSynthesizeHighdimensionalLongitudinal2023},  we assess fidelity using unigram, bigram, and trigram probabilities within each visit, as well as sequential bigram probabilities between consecutive visits. For instance, the bigram probability of the sequence
$[icd_{1053}, icd_{610}]$ is calculated as its frequency divided by the total number of patients. We report Pearson’s correlation between the top-1000 n-gram probabilities of the train and test/synthetic data pairs.

For time series data, we create manual embeddings for each patient by computing common statistics (min, max, mean, std) from the first 48 hours of the patient’s stay. We then report the precision, recall, density, and coverage (PRDC) metrics \cite{naeemReliableFidelityDiversity2020} to evaluate the fidelity between the embeddings of the train split and the test/synthetic split. Additionally, we calculate the pairwise temporal correlation between all time series variables by concatenating the time series data from all patients and report the mean squared error between the correlation matrices of the train split and the test/synthetic split (\(MSE_{corr}\)). We also assess correlation accuracy \cite{liGeneratingSyntheticMixedtype2023} by discretizing the correlation coefficients into five levels: High negative (\([-1, -0.5)\)), Medium negative (\([-0.5, -0.2)\)), Low (\([-0.2, 0.2)\)), Medium positive (\([0.2, 0.5)\)), and High positive (\([0.5, 1)\)). The confusion matrix of these correlation levels is then reported. To assess the fidelity of the missingness patterns, we visualize the co-occurrence of measurements for each pair of time series variables and normalize this by the sum of occurrences for each variable.

\paragraph{Utility}

We assess the utility of the synthetic data by evaluating its performance in two tasks: \textit{in-hospital mortality} prediction (binary classification) and \textit{phenotypes} classification (multi-label classification). Specifically, we use time series embeddings from the first 48 hours of patient admission, as described earlier, and train a LightGBM model \cite{keLightGBMHighlyEfficient2017}. To examine the impact of data augmentation with synthetic data, we incorporate the entire synthetic dataset into the training data at various ratios (0, 0.1, 0.2, 0.5, 1.0) and evaluate the model on the test split.

\paragraph{Privacy}

We adopt the Membership Inference Attack (MIA) approach as described in EHR-Safe \cite{yoonEHRSafeGeneratingHighfidelity2023}, where the goal is to determine whether specific data points were included in the training dataset. To accomplish this, we fit a K-Nearest Neighbors (KNN) model on the synthetic data and calculate the nearest distances for each patient in both the training and test splits. A significant disparity between the distance distributions in the synthetic-train and synthetic-test sets indicates lower privacy. For ICD code sequences, we use the Hamming distance, whereas for time series embeddings, we employ the Euclidean distance. We then fit Gaussian distributions to these distances and assess the differences between the two distributions using the Wasserstein Distance (WD), Jensen-Shannon Divergence (JSD), and Area Under the Receiver Operating Characteristic (AUROC) metrics.




\subsection{Baselines}

For ICD code generation, we compare our method with HALO \cite{theodorouSynthesizeHighdimensionalLongitudinal2023}, a hierarchical autoregressive language model, and PromptEHR \cite{wangPromptEHRConditionalElectronic2022}, which utilizes an encoder-decoder transformer model. For time series generation, we compare our method with RTSGAN \cite{peiGeneratingRealWorldTime2021a} and TimEHR \cite{karamiTimEHRImagebasedTime2024}, both of which are GAN-based models designed for generating irregularly-sampled time series data. Additionally, we include HALO in the comparison for time series generation, as it employs a similar tokenization strategy for numerical variables.

\textbf{Ablation study}. We further demonstrate the effectiveness of numerical tokenization by removing the discretization step and treating the numerical values as raw text. We refer to this model as SynEHRgy0.

\subsection{Training Details}

We employ a small GPT-2 model from the Transformers library\footnote{\href{https://huggingface.co/docs/transformers}{https://huggingface.co/docs/transformers}}. Our model consists of 4 layers, 4 attention heads, and 384 embedding dimensions. After testing various context lengths, we found that 1024 tokens offer a good balance between the quality of generated data and training time (notably, context lengths of 256, 512, 1024, and 2048 correspond to 20\%, 34\%, 51\%, and 68\% of the original data, respectively). For the ablated model (SynEHRgy0), we need to use a larger context length of size 4098 due to its inefficient tokenization.

We use the Adam optimizer with a learning rate of \(3 \times 10^{-4}\) and train the model for 20 epochs on a system equipped with an \textit{NVIDIA Tesla V100-SXM2-32GB} GPU. The batch size is set to 128, and we employ gradient accumulation with a step of two. All other hyperparameters are set to their default values as per the Transformers library. For generation, we use \textit{top-k sampling} with a temperature of 0.7 and a top-k value of 50. We generate 30,000 synthetic patients which is the same size of training split. Due to computational limitation, we have performed minimal hyperparameter tuning for our model and for the baselines to ensure a fair comparison.

\section{Results}

\subsection{Fidelity}

\begin{table}
  \small
  \centering
  \caption{Fidelity metrics for ICD codes}

\begin{tabular}{lccccc}
\toprule
\toprule
      & \multicolumn{3}{c}{single-visit} &       & \multicolumn{1}{c}{\multirow{2}[4]{*}{Sequential bigram}} \\
\cmidrule{2-4}      & unigram & bigram & trigram &       &  \\
\midrule
\textit{Test split} & \textit{0.998} & \textit{0.983} & \textit{0.932} &       & \textit{\textbf{0.954}} \\
\midrule
SynEHRgy & 0.885 & \textbf{0.969} & \textbf{0.916} &       & 0.856 \\
HALO  & \textbf{0.978} & 0.329 & 0.102 &       & \textit{0.839} \\
PromptEHR & 0.747 & 0.246 & 0.114 &       & NA \\
\bottomrule
\bottomrule
\end{tabular}%

  \label{tab:fid-icd}%

\end{table}%

\autoref{tab:fid-icd} shows the correlation values between n-gram probabilities of the training and test/synthetic splits. While HALO achieves the highest performance for unigram probabilities, our method consistently surpasses all baselines, particularly in bigram, trigram, and sequential bigram probabilities. This underscores the ability of our method to generate high-quality synthetic ICD codes.

\begin{table}
  \small
  \centering
  \caption{Fidelity metrics for Time Series. \textbf{Bold} and \underline{underline} values indicate the best and second best results, respectively.}

\begin{tabular}{rccccc}
\toprule
\toprule
      & Precision & Recall & Density & Coverage & MSE\_{corr} \\
\midrule
\textit{Test split} & \textit{0.863 (0.004)} & \textit{0.878 (0.003)} & \textit{0.984 (0.005)} & \textit{0.966 (0.001)} & \textit{0.02} \\
\midrule
SynEHRgy & \textbf{0.797 (0.005)} & \textbf{0.841 (0.004)} & \uline{0.673 (0.01)} & \textbf{0.873 (0.001)} & \textbf{0.036} \\
SynEHRgy0 & 0.452(.018) & 0.227(0.020) & 0.345(0.009) & 0.150(0.014) & 0.133 \\
TimEHR & 0.593 (0.008) & \uline{0.669 (0.011)} & 0.385 (0.011) & \uline{0.613 (0.003)} & \uline{0.059} \\
HALO  & 0.521 (0.021) & 0.469 (0.002) & 0.387 (0.022) & 0.19 (0.003) & 0.072 \\
RTSGAN & \uline{0.725 (0.026)} & 0.113 (0.019) & \textbf{0.744 (0.054)} & 0.108 (0.001) & 0.097 \\
\bottomrule
\bottomrule
\end{tabular}%

  \label{tab:fid-ts}%

\end{table}%

\begin{figure}[h]
  \centering
  \includegraphics[width=0.95\textwidth]{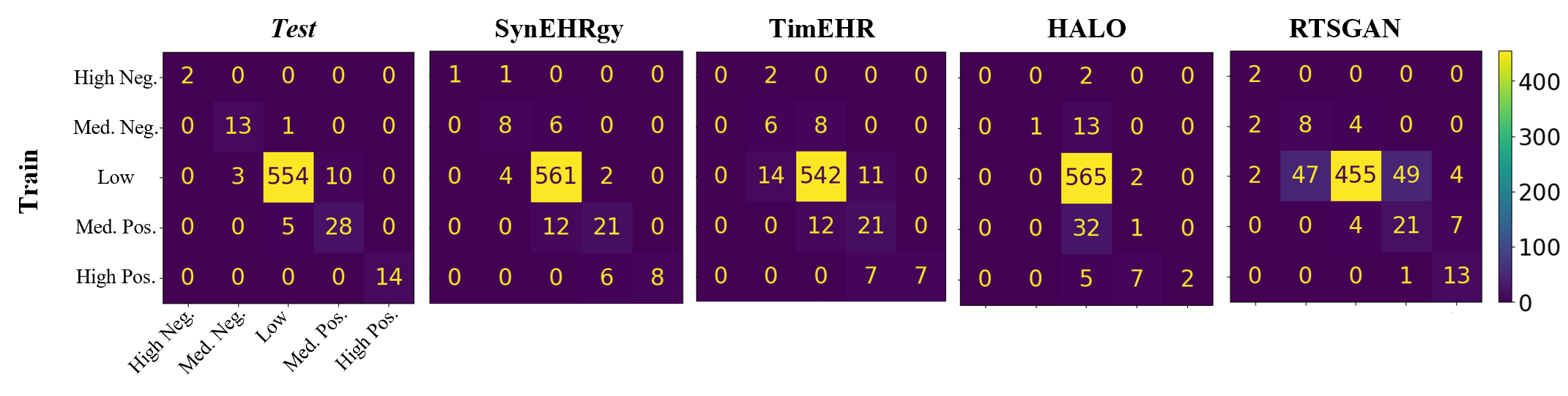}
  \caption{Confusion matrices of correlation levels for time series data}
  \label{fig:confmats}
\end{figure}

\begin{figure}[h]
  \centering
  \includegraphics[width=0.95\textwidth]{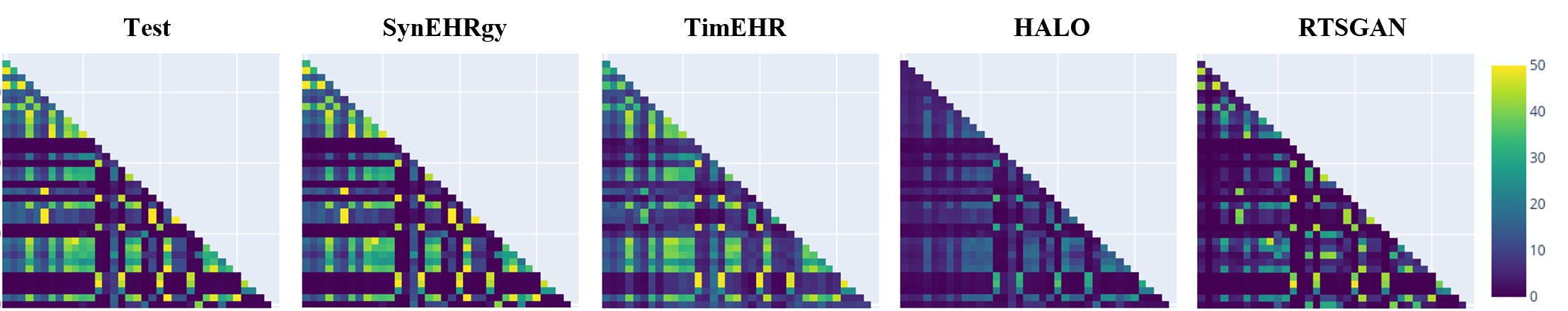}
  \caption{Co-occurrence of measurement for time series data}
  \label{fig:co-occ}
\end{figure}

\autoref{tab:fid-ts} presents the precision, recall, density, and coverage (PRDC) metrics, as well as the temporal correlation difference ($MSE_{corr}$) for evaluating the fidelity of time series embeddings. It is important to consider these PRDC metrics collectively. For instance, while RTSGAN achieves the highest density, it performs poorly on recall and coverage metrics. Our method excels in precision, recall, and coverage, and ranks second in density. Additionally, our model demonstrates the lowest $MSE_{corr}$, indicating strong performance in preserving structural correlations within the data. This is further supported by the confusion matrix of correlation levels shown in \autoref{fig:confmats}, where SynEHRgy displays superior performance in capturing correlation levels. Moreover, the co-occurrence analysis in \autoref{fig:co-occ} reveals that our method exhibits the highest fidelity in preserving missingness patterns, highlighting its capability to generate realistic irregularly-sampled time series data. The ablated model (SynEHRgy0) exhibits poor fidelity performance, demonstrating the importance of numerical tokenization.

\subsection{Utility}

\begin{table}
  \small
  \centering
  \caption{AUROC for phenotypes prediction. \textit{NA}: Not Applicable.}

\begin{tabular}{rccccc}
\toprule
\toprule
      & \multicolumn{5}{c}{Train ratio} \\
\cmidrule{2-6}      & 0 (TSTR) & 0.1   & 0.2   & 0.5   & 1 \\
\midrule
\textit{No synthetic} & \textit{NA} & \textit{0.884} & \textit{0.892} & \textit{0.9} & \textit{0.906 (TRTR)} \\
\textit{Val split} & \textit{0.776} & \textit{0.783} & \textit{0.785} & \textit{0.793} & \textit{0.8} \\
\midrule
SynEHRgy & \textbf{0.775} & \textbf{0.78} & \textbf{0.783} & \textbf{0.791} & \textbf{0.798} \\
TimEHR & 0.731 & 0.756 & 0.763 & 0.771 & 0.779 \\
HALO  & 0.665 & 0.746 & 0.761 & 0.78  & 0.787 \\
RTSGAN & 0.545 & 0.751 & 0.767 & 0.781 & 0.794 \\
\bottomrule
\bottomrule
\end{tabular}%

  \label{tab:util1}%

\end{table}%

\begin{table}
  \small
  \centering
  \caption{AUROC for in-hospital mortality prediction. \textit{NA}: Not Applicable.}

\begin{tabular}{rccccc}
\toprule
\toprule
      & \multicolumn{5}{c}{Train ratio} \\
\cmidrule{2-6}      & 0 (TSTR) & 0.1   & 0.2   & 0.5   & 1 \\
\midrule
\textit{No synthetic} & \textit{NA} & \textit{0.889} & \textit{0.898} & \textit{0.912} & \textit{0.92 (TRTR)} \\
\textit{Val split} & \textit{0.902} & \textit{0.91} & \textit{0.914} & \textit{0.915} & \textit{0.924} \\
\midrule
SynEHRgy & \textbf{0.894} & \textbf{0.902} & \textbf{0.902} & \textbf{0.905} & \textbf{0.915} \\
TimEHR & 0.865 & 0.887 & 0.893 & 0.906 & 0.912 \\
RTSGAN & 0.58  & 0.878 & 0.891 & 0.906 & 0.919 \\
\bottomrule
\bottomrule
\end{tabular}%

  \label{tab:util2}%

\end{table}%

\autoref{tab:util1} and \autoref{tab:util2} present the AUROC scores for the 25-phenotype and in-hospital mortality prediction tasks, respectively, across different data augmentation settings. When training exclusively on synthetic data and testing on real data (TSTR), our method surpasses all baseline models and achieves performance closest to that of the Train on Real and Test on Real setting (TRTR). Additionally, our method demonstrates the highest AUROC when synthetic data is incorporated into the training set, although the performance gain diminishes as the proportion of real training data increases. This suggests that while the models leverage the synthetic data effectively, they do not extend beyond the real data distribution to generate new patient profiles that significantly enhance test performance. Furthermore, the validation split shows marginally better performance compared to other synthetic data settings. These utility results highlight the potential of our method for data anonymization (with no exposure of training data) and for data augmentation in scenarios with limited training data.

\subsection{Privacy}

\begin{table*}
  \small
  \centering
  \caption{Privacy metrics for ICD codes and Time Series}

\begin{tabular}{ccccccccc}
\toprule
\toprule
      & \multicolumn{3}{c}{ICD codes} &       &       & \multicolumn{3}{c}{Time Series} \\
\cmidrule{1-4}\cmidrule{6-9}Model & JSD   & WD    & AUROC &       & Model & WD    & JSD   & AUROC \\
\cmidrule{1-4}\cmidrule{6-9}SynEHRgy & 0.012 & 0.000 & 0.510 &       & SynEHRgy & 0.003 & 0.001 & 0.494 \\
HALO  & 0.013 & 0.000 & 0.511 &       & TimEHR & 0.003 & 0.001 & 0.492 \\
PromptEHR & 0.012 & 0.001 & 0.455 &       & HALO  & 0.002 & 0.001 & 0.493 \\
      &       &       &       &       & RTSGAN & 0.002 & 0.001 & 0.494 \\
\bottomrule
\bottomrule
\end{tabular}%

  \label{tab:privacy}%

\end{table*}%

Membership Inference Attack (MIA) metrics are reported in \autoref{tab:privacy}. As can be seen, none of methods shows any potential privacy risk. It should be noted that these privacy metrics are computed based on hamming distance of ICD codes and Euclidean distance of time series data, and the results may vary for different distance metrics.

\section{Conclusion}

In this work, we introduced SynEHRgy, a novel methodology for generating synthetic structured EHR data that incorporates mixed-type data (ICD codes and time series) across multiple visits. A key innovation of our approach is the tokenization strategy for numerical variables, which enhances compatibility with decoder-only transformer models. Despite its advantages, the proposed method has limitations. Specifically, treating each data point (e.g., an ICD code or a heart rate measurement) as a single token may become inefficient for very large sequences due to context length constraints. Future work will explore more efficient tokenization strategies to mitigate this issue. Additionally, integrating high-frequency continuous signals, such as ECG data, is currently not feasible with our tokenization approach. We plan to address this limitation by incorporating patch-based tokenization strategies \cite{wooUnifiedTrainingUniversal2024, ansariChronosLearningLanguage2024, liuTimerGenerativePretrained2024}. Furthermore, we aim to expand our methodology to include other data modalities, such as clinical notes and radiology images, to generate comprehensive multimodal EHR data. Another promising direction for future research is to enhance the capabilities of clinical LLMs, such as Meditron \cite{chenMEDITRON70BScalingMedical2023} and Med-PaLM \cite{singhalLargeLanguageModels2022}, to better understand and generate structured data.

  {
    \small

    \printbibliography

  }






\newpage
\section*{NeurIPS Paper Checklist}

The checklist is designed to encourage best practices for responsible machine learning research, addressing issues of reproducibility, transparency, research ethics, and societal impact. Do not remove the checklist: {\bf The papers not including the checklist will be desk rejected.} The checklist should follow the references and follow the (optional) supplemental material.  The checklist does NOT count towards the page
limit.

Please read the checklist guidelines carefully for information on how to answer these questions. For each question in the checklist:
\begin{itemize}
  \item You should answer \answerYes{}, \answerNo{}, or \answerNA{}.
  \item \answerNA{} means either that the question is Not Applicable for that particular paper or the relevant information is Not Available.
  \item Please provide a short (1–2 sentence) justification right after your answer (even for NA).
\end{itemize}

{\bf The checklist answers are an integral part of your paper submission.} They are visible to the reviewers, area chairs, senior area chairs, and ethics reviewers. You will be asked to also include it (after eventual revisions) with the final version of your paper, and its final version will be published with the paper.

The reviewers of your paper will be asked to use the checklist as one of the factors in their evaluation. While "\answerYes{}" is generally preferable to "\answerNo{}", it is perfectly acceptable to answer "\answerNo{}" provided a proper justification is given (e.g., "error bars are not reported because it would be too computationally expensive" or "we were unable to find the license for the dataset we used"). In general, answering "\answerNo{}" or "\answerNA{}" is not grounds for rejection. While the questions are phrased in a binary way, we acknowledge that the true answer is often more nuanced, so please just use your best judgment and write a justification to elaborate. All supporting evidence can appear either in the main paper or the supplemental material, provided in appendix. If you answer \answerYes{} to a question, in the justification please point to the section(s) where related material for the question can be found.

IMPORTANT, please:
\begin{itemize}
  \item {\bf Delete this instruction block, but keep the section heading ``NeurIPS paper checklist"},
  \item  {\bf Keep the checklist subsection headings, questions/answers and guidelines below.}
  \item {\bf Do not modify the questions and only use the provided macros for your answers}.
\end{itemize}


\begin{enumerate}

  \item {\bf Claims}
  \item[] Question: Do the main claims made in the abstract and introduction accurately reflect the paper's contributions and scope?
  \item[] Answer: \answerYes{} 
  \item[] Justification: Our work is mostly empirical rather than theoretical. We have deeply evaluated the quality of synthetic data in terms of utility, fidelity and privacy and compared it with the state-of-the-art models.

  \item {\bf Limitations}
  \item[] Question: Does the paper discuss the limitations of the work performed by the authors?
  \item[] Answer: \answerYes{} 
  \item[] Justification: In the conclusion section, we discussed the limitations of our work such as the tokenization strategy might not be efficient for very large sequences due to limited context length. Our work is tested on a real-world dataset and the results may vary for different datasets. We are open to discussing more limitations in the paper.

  \item {\bf Theory Assumptions and Proofs}
  \item[] Question: For each theoretical result, does the paper provide the full set of assumptions and a complete (and correct) proof?
  \item[] Answer: \answerNA{} 
  \item[] Justification: Our work is mostly empirical and does not include theoretical results.

  \item {\bf Experimental Result Reproducibility}
  \item[] Question: Does the paper fully disclose all the information needed to reproduce the main experimental results of the paper to the extent that it affects the main claims and/or conclusions of the paper (regardless of whether the code and data are provided or not)?
  \item[] Answer: \answerYes{} 
  \item[] Justification: The dataset is publicly available for researchers. The model architecture and training details are provided in the main paper. Our work is highly replicable as we use the Transformer pipeline which is widely used in the community.

  \item {\bf Open access to data and code}
  \item[] Question: Does the paper provide open access to the data and code, with sufficient instructions to faithfully reproduce the main experimental results, as described in supplemental material?
  \item[] Answer: \answerYes{} 
  \item[] Justification: The dataset (MIMIC-III) is publicly available. We will release the code and preprocessing pipeline upon acceptance.
  \item[] Guidelines:
    \begin{itemize}
      \item The answer NA means that paper does not include experiments requiring code.
      \item Please see the NeurIPS code and data submission guidelines (\url{https://nips.cc/public/guides/CodeSubmissionPolicy}) for more details.
      \item While we encourage the release of code and data, we understand that this might not be possible, so “No” is an acceptable answer. Papers cannot be rejected simply for not including code, unless this is central to the contribution (e.g., for a new open-source benchmark).
      \item The instructions should contain the exact command and environment needed to run to reproduce the results. See the NeurIPS code and data submission guidelines (\url{https://nips.cc/public/guides/CodeSubmissionPolicy}) for more details.
      \item The authors should provide instructions on data access and preparation, including how to access the raw data, preprocessed data, intermediate data, and generated data, etc.
      \item The authors should provide scripts to reproduce all experimental results for the new proposed method and baselines. If only a subset of experiments is reproducible, they should state which ones are omitted from the script and why.
      \item At submission time, to preserve anonymity, the authors should release anonymized versions (if applicable).
      \item Providing as much information as possible in supplemental material (appended to the paper) is recommended, but including URLs to data and code is permitted.
    \end{itemize}

  \item {\bf Experimental Setting/Details}
  \item[] Question: Does the paper specify all the training and test details (e.g., data splits, hyperparameters, how they were chosen, type of optimizer, etc.) necessary to understand the results?
  \item[] Answer: \answerYes{} 
  \item[] Justification: We have used a famous preprocessing pipeline for MIMIC-III dataset which splits the data into train, validation and test sets. The labels for downstream tasks are also selected using the same pipeline. We have provided the non-default hyperparameters used for training the model using Huggingface's Transformers library.

  \item {\bf Experiment Statistical Significance}
  \item[] Question: Does the paper report error bars suitably and correctly defined or other appropriate information about the statistical significance of the experiments?
  \item[] Answer: \answerNo{} 
  \item[] Justification: As our training split is large enough (~30k), we haven't performed k-fold evaluation because of computational constraints. Yet, there are some ways to report error bars such as running the training procedure multiple times with different random seeds and reporting the mean and standard deviation of the results.

  \item {\bf Experiments Compute Resources}
  \item[] Question: For each experiment, does the paper provide sufficient information on the computer resources (type of compute workers, memory, time of execution) needed to reproduce the experiments?
  \item[] Answer: \answerYes{} 
  \item[] Justification: This information is provided in the training details section of the paper.

  \item {\bf Code Of Ethics}
  \item[] Question: Does the research conducted in the paper conform, in every respect, with the NeurIPS Code of Ethics \url{https://neurips.cc/public/EthicsGuidelines}?
  \item[] Answer: \answerYes{} 
  \item[] Justification: Our dataset is publicly available and we have used a widely used model architecture for our experiments. We have tried our best to preserve the anonymity.

  \item {\bf Broader Impacts}
  \item[] Question: Does the paper discuss both potential positive societal impacts and negative societal impacts of the work performed?
  \item[] Answer: \answerYes{} 
  \item[] Justification: The development of SynEHRgy, while offering significant advances in the generation of synthetic EHR data, comes with broader ethical and societal implications that must be carefully considered. In terms of safety, while the synthetic data generated by our method is not linked to real individuals, there remains the potential risk of misuse in creating misleading datasets that could negatively impact clinical decision-making or research outcomes. Security risks are mitigated by our focus on synthetic data, which reduces the likelihood of exposing real patient information; however, the generated data could still be vulnerable to manipulation or malicious use in healthcare applications if not properly safeguarded. Regarding discrimination, care must be taken to ensure that biases in the original datasets, such as underrepresentation of certain demographic groups, do not carry over into synthetic data and perpetuate disparities in healthcare outcomes. Furthermore, while SynEHRgy does not involve surveillance directly, we recognize that synthetic data might be used to develop surveillance systems or predictive tools that could target vulnerable populations. We strongly advocate against any use of our method that could lead to deceptive practices, including fraud or impersonation, and emphasize the need for transparency in how the data is applied. From an environmental perspective, training large-scale models like transformers requires substantial computational resources, and future work should explore ways to reduce the environmental footprint of these models. Finally, to address fairness and bias, we recommend rigorous evaluation of the synthetic data to identify and correct any imbalances or biases inherited from the original dataset, ensuring equitable outcomes in downstream applications.

  \item {\bf Safeguards}
  \item[] Question: Does the paper describe safeguards that have been put in place for responsible release of data or models that have a high risk for misuse (e.g., pretrained language models, image generators, or scraped datasets)?
  \item[] Answer: \answerNA{} 
  \item[] Justification: \answerNA{}

  \item {\bf Licenses for existing assets}
  \item[] Question: Are the creators or original owners of assets (e.g., code, data, models), used in the paper, properly credited and are the license and terms of use explicitly mentioned and properly respected?
  \item[] Answer: \answerNA{} 
  \item[] Justification: \answerNA{}

  \item {\bf New Assets}
  \item[] Question: Are new assets introduced in the paper well documented and is the documentation provided alongside the assets?
  \item[] Answer: \answerNA{} 
  \item[] Justification: \answerNA{}

  \item {\bf Crowdsourcing and Research with Human Subjects}
  \item[] Question: For crowdsourcing experiments and research with human subjects, does the paper include the full text of instructions given to participants and screenshots, if applicable, as well as details about compensation (if any)?
  \item[] Answer: \answerNA{} 
  \item[] Justification: \answerNA{}

  \item {\bf Institutional Review Board (IRB) Approvals or Equivalent for Research with Human Subjects}
  \item[] Question: Does the paper describe potential risks incurred by study participants, whether such risks were disclosed to the subjects, and whether Institutional Review Board (IRB) approvals (or an equivalent approval/review based on the requirements of your country or institution) were obtained?
  \item[] Answer: \answerNA{} 
  \item[] Justification: \answerNA{}

\end{enumerate}

\end{document}